\documentclass[sigconf, authordraft,review=false]{acmart}
\renewcommand\footnotetextcopyrightpermission[1]{} 
\makeatletter
\renewcommand\@formatdoi[1]{\ignorespaces}
\makeatother
\usepackage{booktabs} 
\usepackage{xcolor}

\acmDOI{10.475/123_4}

\acmISBN{123-4567-24-567/08/06}

\begin{document}
\title{Video-based Person Re-identification Using Spatial-Temporal Attention Networks}

\author{Shivansh Rao}
\orcid{1234-5678-9012}
\affiliation{%
  \institution{Delhi Technological University}
}
\email{shivansh_bt2k15@dtu.ac.in}

\author{Tanzila Rahman}
\orcid{1234-5678-9012}
\affiliation{%
  \institution{University of Manitoba}
}
\email{rahmant4@cs.umanitoba.ca}

\author{Mrigank Rochan}
\orcid{1234-5678-9012}
\affiliation{%
  \institution{University of Manitoba}
}
\email{mrochan@cs.umanitoba.ca}

\author{Yang Wang}
\orcid{1234-5678-9012}
\affiliation{%
  \institution{University of Manitoba}
}
\email{ywang@cs.umanitoba.ca}

\begin{abstract}
We consider the problem of video-based person re-identification. The goal is to identify a person from videos captured under different cameras. In this paper, we propose an efficient spatial-temporal attention based model for person re-identification from  videos. Our  method  generates  an  attention  score for each frame based on frame-level features. The attention scores of all frames in a video are used to produce a weighted feature vector for the input video. Unlike most existing deep learning methods that use global representation, our approach focuses on attention scores. Extensive experiments on two benchmark datasets demonstrate that our method achieves the state-of-the-art performance. Code is available at \textit{\textcolor{red}{\url{https://github.com/rshivansh/Spatial-Temporal-attention}}}.
\end{abstract}

%
%

\ccsdesc[500]{Computing methodologies~Computer vision}
\ccsdesc[500]{Computing methodologies~Object identification}

\keywords{Person Re-Identification, Attention Mechanism }

\maketitle

\begin{figure*}
\includegraphics[width = 1.0 \linewidth]{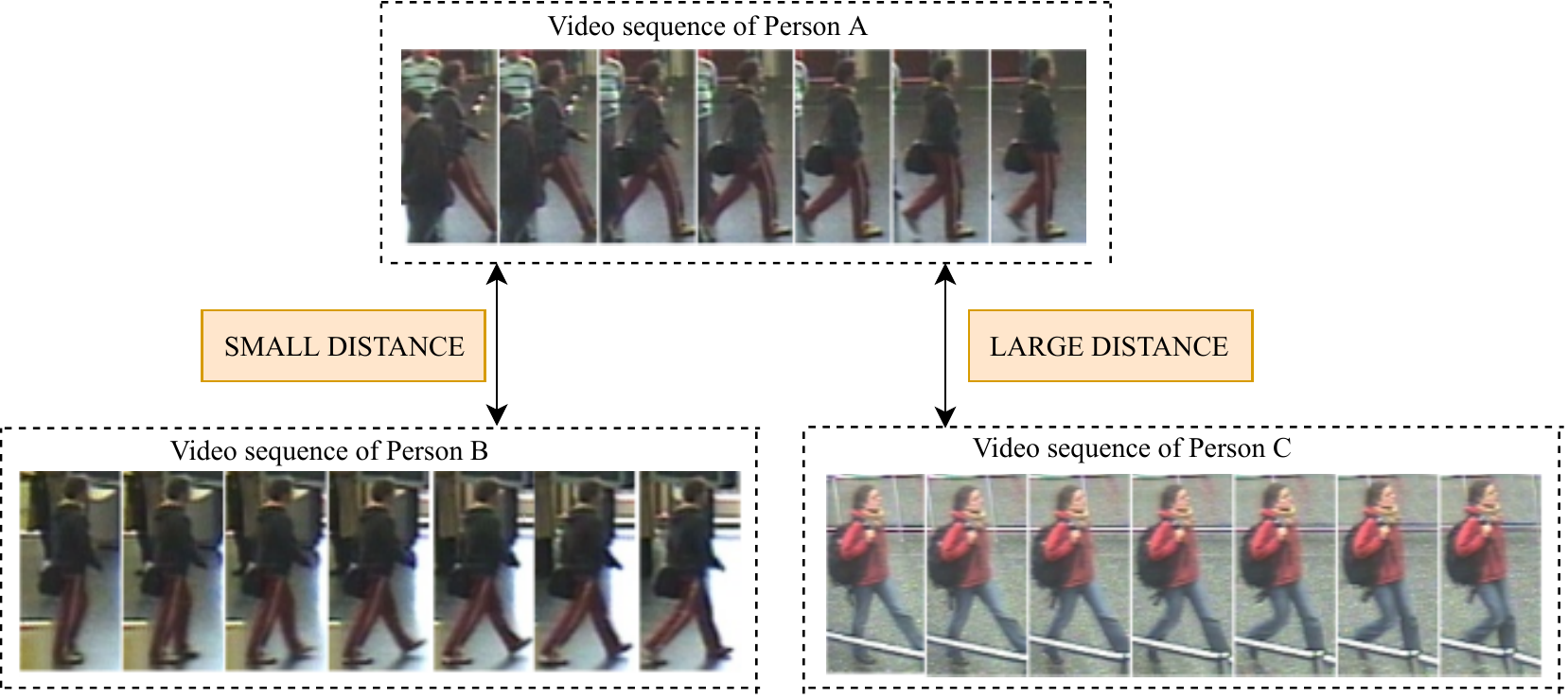}
\caption{Illustration of the video-based person re-identification problem.  In this case, our goal is to identify person A from two video
sequences in the second row.  If two videos contain the same person, we would like the distance between them to be small.  Otherwise,
we would like the distance to be large. Some frames in a video sequence may be affected by occlusions and are not informative about the
person\textup{'}s identity. In this paper, we use an attention model to focus on informative frames for re-identification.}
\end{figure*}
\section{Introduction}
In this paper, our goal is to solve the problem of video-based person re-identification.  Given a video containing a person, the goal is to identify the same person from other videos possibly captured under different cameras.  Person re-identification is useful in a wide range of applications,
e.g video surveillance, police investigation, etc. A common strategy for person re-identification is to formulate it as a metric learning problem.  Given the query video and a candidate video, the goal is to develop algorithms to compute the distance between these two videos.   If the distance is small, it means the two videos likely contain the same person. See Figure 1 for an illustration. Previous work in person re-identification falls into two broad categories: image-based re-identification and video-
based re-identification. Earlier work ~\cite{liao15_iccv,qian17_iccv,ustinova17_avss,varior16_eccv,
xiao16_cvpr,yi14_icpr,zhangL16_cvpr,zhengL15_iccv} in this area focuses on the former, where the inputs to these systems are pairs of images and the goal is  to  identify  whether  they  are  images  of  the  same  person.   Recently,  video-based  person  re-identification  is  receiving  increasing  attention ~\cite{li17_workshop,liu15_iccv,mclaughlin16_cvpr,wang14_eccv,
xu17_iccv,zhou17_cvpr,zhu16_ijcai}. Compared with static images, video-based person re-identification is a more natural setting for practical applications such as video surveillance. 

Person re-identification (either image or video based) is a challenging problem since the images/videos are often captured under different camera views. This can cause large variations in illumination, body pose, viewpoint, etc. Compared with static images, the temporal information in videos can potentially provide additional information that can help
disambiguate the identity of a person. Previous work in this area has explored ways of exploiting this temporal information. A common strategy ~\cite{mclaughlin16_cvpr,xu17_iccv,zhou17_cvpr} is to use temporal pooling to combine frame-level features to represent the entire video sequence.  Then this video-level feature vector
can be used for re-identification.

Previous work ~\cite{xu17_iccv,zhou17_cvpr} has made the observation that not all frames in a video are informative. For example, if the person is occluded in a frame, ideally we would like the feature representation of the video to ignore this frame and focus on other useful frames. A natural way of solving this problem is to use the attention models  ~\cite{bahdanau15_iclr,shih16_cvpr,xu15_icml} that  have  been  popular  in  visual  recognition  recently. In ~\cite{xu17_iccv,zhou17_cvpr}
, RNN is used to model the temporal information of the frames and generate the attention score for each frame for person re-identification.

In  this  paper,  we  propose  a  new  attention  model  for video-based  re-identification.
Compared  with  previous works ~\cite{xu17_iccv,zhou17_cvpr}, our model has several novelties.  First, instead of using RNN, we directly produce the attention score of each frame based on the image feature of this frame. Our experimental results show that this simpler method outperforms  RNN-based  attention  method.   Since  the  attention score of each frame is calculated based on the frame,  the computation of attention scores over all frames can be easily made parallel and take full advantage of the GPU hardware. Second we propose a new spatio-temporal attention model that allows useful information to be extracted from each frame without succumbing to occlusions and misalignments.

Our contributions include:

1.  A new attention mechanism for video-based person  re-
identification. Unlike  previous  work ~\cite{zhou17_cvpr}  that uses  RNN  to  generate  the  attentions,  our  model  directly generates attentions based on frame-based features. As a consequence, the computation of the attentions is much simpler and can be easily parallelized.
In contrast, RNN has to process frames in a sequential order, so the computation cannot be made parallel. Despite of its simplicity, our model outperforms the more sophisticated RNN-based attention mechanism in ~\cite{zhou17_cvpr} .\\

2. Our newly designed network learns multiple spatial attention along with temporal attention. In order to represent the overall 
spatial attention of the frame, we need multiple hops of attention ~\cite{lin2017structured} that focus on different parts of the frame. In addition, we also study the effect of incrementing the value of number of hops of spatial attention.

\section{Related Work}
There    has    been    extensive    work    on    person    re- identification from static images.   Early work in this area uses hand-crafted feature representations ~\cite{gray08_eccv,liao15_cvpr,ma12_workshop,matsukawa16_cvpr,zhao14_cvpr}. Most  of  these  methods  involve  extracting  feature  representations  that  are  invariant  to  viewpoint  changes,  then
learning a distance metric to measure the similarity of two images.

Deep  learning  approaches,  in  particularly  deep  convolutional  neural  networks  (CNNs),  have  achieved  tremendous successes in various visual recognition tasks ~\cite{krizhevsky12_nips}.  In many areas of computer vision, CNNs have replaced hand-engineering  feature  representations  with  features  learned end-to-end from data.  Recently, CNNs have been used for image-based  person  re-identification ~\cite{liao15_iccv,qian17_iccv,ustinova17_avss,varior16_eccv,xiao16_cvpr,yi14_icpr,
zhangL16_cvpr,zhengL15_iccv}.   These  methods  use  deep  network  architecture  such  as  Siamese  network  ~\cite{hadsell06_cvpr}  to  map  images  to  feature vectors.  These feature vectors can then be used for re- identification.   Although  the  performance  of  image-based person  re-identification  has  increased  significantly,  this  is not a very realistic setting for practical applications.
\begin{figure*}
\includegraphics[width = 1.0 \linewidth]{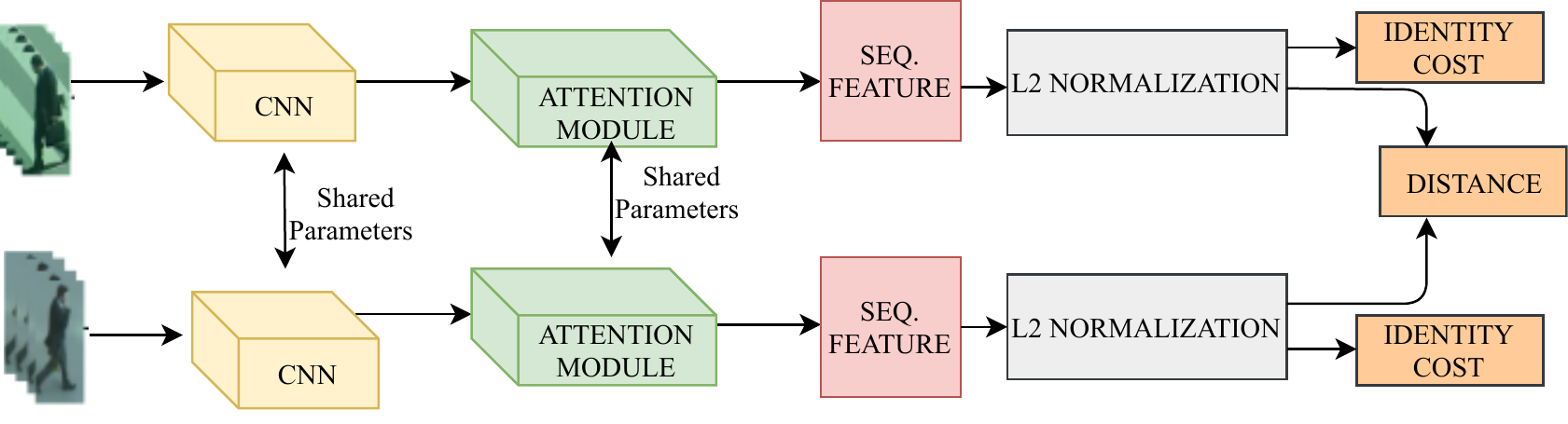}
\caption{Overall architecture of our proposed Siamese network. It takes two input video sequences and pass to the Convolutional Neural
Network (CNN) to extract features on each frame. The output from the CNN is fed to the attention module and generate an attention score
for each frame.  These attention scores combined with frame-level feature vectors to form a feature vector for the
whole video. The video-level feature vectors are compared to decide whether the videos contain the same person.}
\end{figure*}
To address the limitation of image-based re-identification,  a  lot  of  recent  work  has  began  to  explore
video-based re-identification ~\cite{li17_workshop,liu15_iccv,mclaughlin16_cvpr,
wang14_eccv,xu17_iccv,yan16_eccv,zhou17_cvpr,zhu16_ijcai} since   it   is   closer   to   real-world   application   settings. Compared  with  static  images,   videos  contain  temporal information  that  is  potentially  distinctive  for  differentiating  a  person\textup{'}s  identity.   Some  prior  work  has  explored
ways   of   incorporating   temporal   information   in   deep convolutional  neural  network  for  re-identification. For example, McLaughlin et al . ~\cite{mclaughlin16_cvpr} use CNN on each frame in  a  video  and  incorporate  a  recurrent  layer  on  the  CNN features. Temporal  pooling  is  then  used  to  combine frame-level features into a single video-level feature vector for re-identification.

Our work is also related to a line of research on incorporating attention mechanism in deep neural networks. The
attention mechanism allows the neural networks to focus on
part of the input and ignore the irrelevant information. It has
been successfully used in many applications, including machine translation [1], image captioning ~\cite{xu15_icml}, visual question
answering ~\cite{shih16_cvpr}, etc. In video-based re-identification, the attention  mechanism  has  also  been  explored ~\cite{xu17_iccv,zhou17_cvpr} . The
intuition is that only a small portion of the video contains
informative information for re-identification.  So the attention mechanism can be used to help the model focus on the informative part of the video.

The work in ~\cite{zhou17_cvpr} is the closest to ours. It uses an RNN to
generate temporal attentions over frames, so that the model
can focus on the most discriminative frames in a video for
re-identification.  In this paper, we uses temporal attentions
over frames as well. But instead of using RNN-based models to generate attentions ~\cite{zhou17_cvpr}, we directly calculate the attention scores based on frame-based features.  This makes
the model much simpler and the computation of attention
scores can be easily parallelized over frames.   We also propose propose a new spatio-temporal attention model wherein we calculate multiple spatial attention over the frames .  We demonstrate that this multiple spatial attention mechanism improves the performance of the final model.

\section{Our Approach}
Figure 2 shows the overall architecture of our proposed
approach based on the Siamese network ~\cite{hadsell06_cvpr}.  The input to
the  Siamese  network  is  a  pair  of  video  sequences  corresponding to the query video and the candidate video to be
compared.  The output of the Siamese network is a scalar
value  indicating  how  likely  these  two  videos  contain  the
same person. Each video goes into one of the two branches
of the Siamese network.  Each branch of the Siamese net-
work is a Convolutional neural network used to extract the
features of the input video. The parameters of two branches
of  the  Siamese  network  are  shared.   Finally,  the  features
from the two input videos are compared to produce the final
output.

When a video goes through one of the two branches of
the Siamese network, we first extract per-frame features on
each  frame  of  the  input  video.   Then  we  compute  an  attention score on each frame indicating how important this
frame is for the re-identification task.  The intuition is that
not  all  frames  in  a  video  are  informative.   The  attention
scores enable our model to ignore certain frames and only
pay attention to informative frames in the video.   The attention scores are then used to aggregate per-frame visual
features weighted by the corresponding attention score to
form a feature vector for the entire video sequence. Along with calculation of temporal attention we also calculate the spatial attention for multiple hops over the input frame. Here the intuition is that in order to pay attention to multiple parts of the frame we need multiple attentions over it, which now allows useful information to be extracted without being hindered by occlusions and misalignments.  We can repeat this process for several hops of attention (see Sec.   4.4),  where each hop
produces attention scores that focus more on the informative frames.  Finally,  the features of two input videos are
compared to produce the output.
\subsection{ Frame-Level Features}
Similar  to  ~\cite{mclaughlin16_cvpr} ,  we  extract  frame-level  features  using
both RGB color and optical flow channels. The colors contain information about the appearance of a person, while the
optical  flows  contain  information  about  the  movement  of
the person.  Intuitively, both of them are useful to differentiate the identity of the person. As a preprocessing step, we
convert all the input images (i.e.  video frames) from RGB
to YUV color space.  We normalize each color channel to
have a zero mean and unit variance. The Lucas-Kanade algorithm ~\cite{lucas1981_iterative} is used to calculate both vertical and horizontal
optical flow channels on each frame. We resize each frame
to have a spatial dimension of 56 $\times$ 40.  The optical flow
field F of the frame is split into two scalar fields $F_x$ and $F_y$
corresponding to the x and y components of the optical flow. In the end, each frame is represented as a 56  $\times$ 40  $\times$ 5 input, where the 5 channels correspond to 3 color channels (RGB) and 2 optical flow channels (x and y).  We fine-tuned CNN architecture of ~\cite{mclaughlin16_cvpr} to extract frame-level features for an input video. 
\begin{figure}[h]
\includegraphics[width = 1.0 \linewidth]{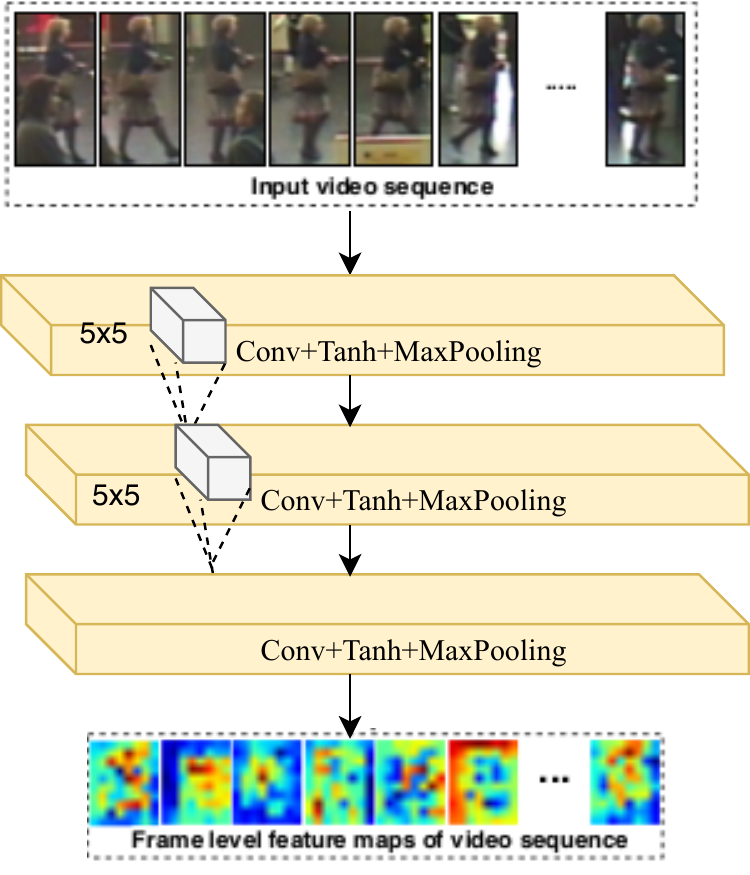}
\caption{Our CNN architecture for extracting frame-level features.  The  network  processes  a  frame  (both  color  and  optical  flow channels)  using  a  series  of  layers }
\end{figure}
The CNN architecture (shown in Fig. 3) consists of three stages of convolution, max-pooling, and nonlinear (tanh) activations. Each convolution filter uses 5  $\times$ 5 kernels with 1  $\times$ 1
stride and 4  $\times$ 4 zero padding. A fully connected layer is used in the end to produce a 128-dimensional feature vector.   Given
T frames in an input video,  the CNN model is  applied  on  each  input  frame  of  the video.  In the end, CNN produces a 128-dimensional feature  vector   (i.e. $x_i$ \(\in\)  \(\mathbb{R}\)$^{\text{128}}$) to represent each frame $x_i$
(i= 1,2,...,T) in the input video.

\subsection{Temporal Attention Network}
Motivated by the recent success of attention based models ~\cite{bahdanau15_iclr,santos16_arxiv,xu15_icml,yin16_tacl}, we propose an attention based approach
for re-identifying person from videos. The intuition behind
the attention based approach is inspired by the human visual
processing ~\cite{xu17_iccv}. Human brains often pay attention to different regions of different sequences when trying to re-identify
persons from videos. Based on this intuition, we propose a
deep Siamese architecture where each branch generates attention scores of different frames based on the frame-level
CNN features.  The attention score of a frame indicates the
importance of this frame for the re-identification task. 

As  shown in  Figure  2,  each  input video  sequence  (sequence of frames with optical flow) is passed to the CNN to extract frame-level feature maps.  Using fully connected
layers, CNN generates feature vector for each video frame.
The sequence of feature vectors are passed to the attention
network to generate attention score.  More specifically, for
each feature vector $x_i$ corresponding to the \textit{i}-th frame, we compute an attention score \(\alpha\)$^t_i$ indicating the importance of this frame.   The attention score is obtained by applying a linear mapping followed by a sigmoid function.   Here,we
use the same parameters for the linear mapping on all
frames.   Let \(\theta\) be  the  vector  of  parameters  for  the  linear mapping. Now  the  attention  score \(\alpha\)$_i^t$  is  calculated  using the following equations:
\begin{equation}
\alpha_i^t=\sigma(\theta^Tx_i), \hspace{0.5cm}i=1,\cdots,T
\end{equation}

Where $\sigma(\cdot)$ is the sigmoid function and $\alpha^t$ is the temporal attention matrix .We  have  also  tried  using  softmax  instead  of  sigmoid
function in Eq. 1 and found that it does not perform as good
as the sigmoid function. Previous work ~\cite{zhao2017_deeply} has made similar observations. Once we have obtained an attention score \(\alpha\)$_i^t$
for each frame in the video,  we can then combine the
attention scores \(\alpha\)$_i^t$(i = 1,2,...,T)  with frame-level feature
vectors to create a weighted feature vector
$f^t$
as follows:
\begin{equation}
f^t=\sum_{i=1}^{T} \alpha^t_ix_i
\end{equation}
where
$f^t$
can be seen as a temporal feature vector for the entire video
which takes into account the importance of each frame in
the video.
\begin{figure*}
\includegraphics[width = 1.0 \linewidth]{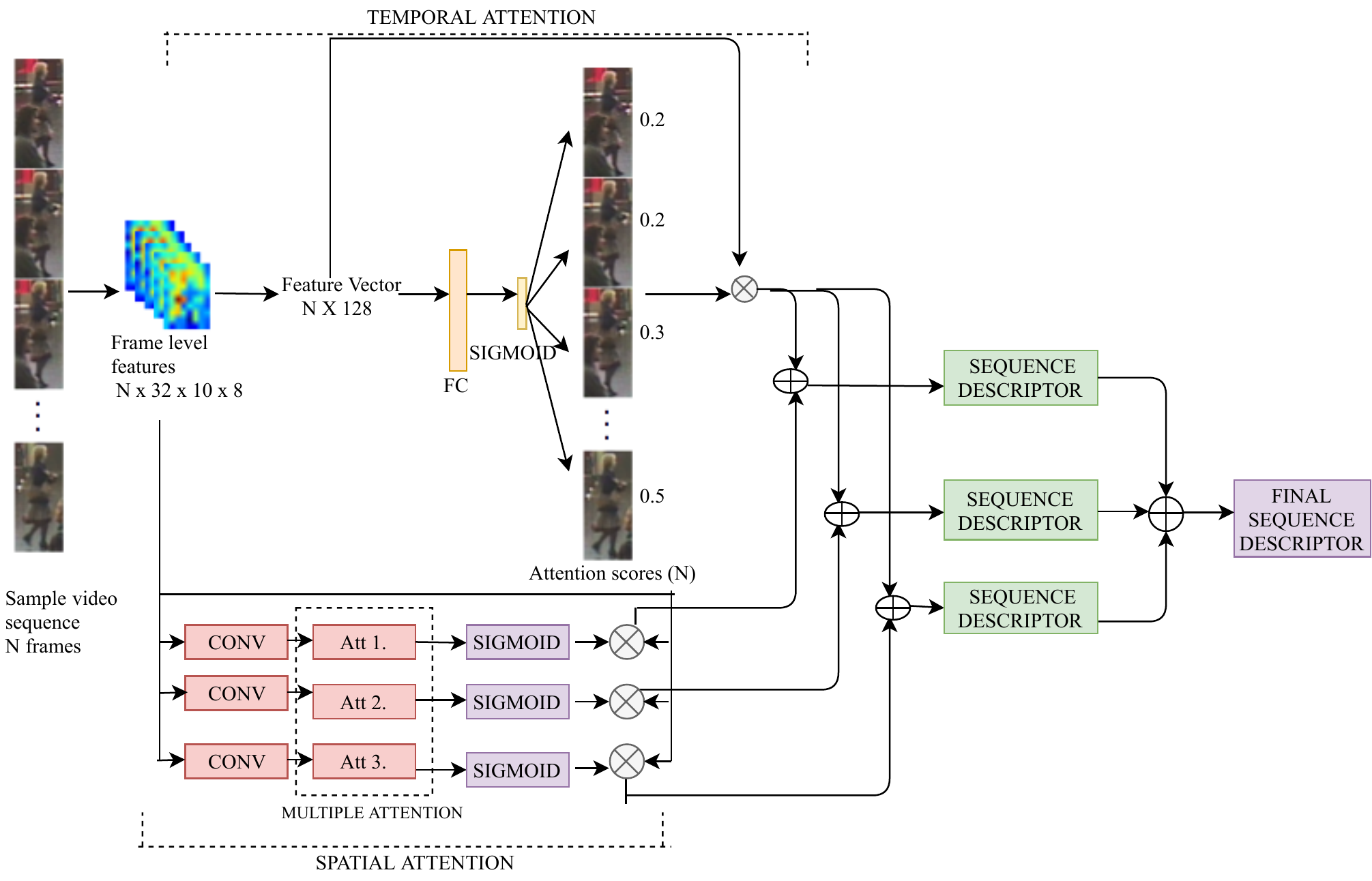}
\caption{Illustration of our proposed attention network architecture. Attention module consists of 2 branches, in the temporal branch (upper) we generate N attention scores by applying linear mapping on the feature vectors followed by a sigmoid function. In the lower branch we calculate multiple spatial attention over the frame level features followed by a sigmoid function, spatial attention features are then added with temporal attention features to form final attention scores for the entire video.}
\end{figure*}
\subsection{Spatial Attention Network}

For the task of  person  re-identification,  due  to  several real time issues such as the  overlooking  angle of most of the surveillance equipments, the pedestrians who are captured only take
a part in whole spatial images.  Therefore, local spatial attention is necessary for deep networks such that useful information is extracted from each frame without getting hindered by problems such as occlusions and misalignments, and thus in this way superfluous information can be removed. In addition we employ multiple spatial attention model which can help us pay attention to multiple parts of the frame by hopping multiple times over it. In the
experiment section, we will show that this multiple attention mechanism improves the performance of our model.

The basic idea of our spatial attention network is to pass the frame level features as input to the spatial attention network which has a convolution layer followed by sigmoid . Note that since our method evaluates multiple attention (see Fig.4), we will get j (where j is number of attention layers) feature vectors from the spatial attention network. The convolution filter ,$conv_j(\cdot)$, uses
5 $\times$ 5 kernels with 1 $\times$  1 stride and 2 $\times$  2 padding. $\alpha_i^{s_j}$ is the attention matrix which is multiplied by the frame level features $x_i$ to get the weighted attention features as follows:
\begin{equation}
\alpha_i^{s_j}=\sigma(conv(x_i)), \hspace{0.2cm}i=1,\cdots,T
\end{equation}
\begin{equation}
f^{s_j}=\sum_{i=1}^{T} \alpha^{s_j}_ix_i, \hspace{0.6cm}i=1,\cdots,T
\end{equation}
 
where $f^{s_j}$ is the spatial attention feature vector of the input video. Now we add spatial attention feature vector $f^{s_j}$ and temporal attention feature vector $f^t$ and then sum it up for all attention layers to get the final feature vector as follows:
\begin{equation}
F_j = f^{s_j} + f^t
\end{equation}
\begin{equation}
F = \sum_j F_j
\end{equation}
We find out empirically the value of j to be 3 ( See section 4.4 for more details).
\subsection{Model Learning}
Our model is a form of the Siamese network (Figure 2).
It has two identical branches with shared parameters.  The
detail  architecture  of  each  branch  is  shown  in  Figure  4.  Let $\textit{F}_1$ and  $\textit{F}_2$ be the feature vectors
of two input videos from the Siamese network. We now calculate Euclidean distance between the feature vectors in a manner similar to ~\cite{mclaughlin16_cvpr,xu17_iccv} and apply the squared hinge loss (Loss$_\textit{hinge}$) as follows:\newline

\hspace{1cm}\(\mathbb{L}\)$_\textit{hinge}$= 
\(\begin{cases}
1/2\left\|  F_1 - F_2 \right\|^2, & X_1 = X_2\\
1/2\lceil max(0,m-\left\|  F_1 - F_2  \right\|) \rceil^2, & X_1 \neq X_2
\end{cases}\) \hspace{0.3cm}(7)\newline

where the hyper-parameter m represents the margin of separating two classes in \(\mathbb{L}\)$_\textit{hinge}$. Here we use $X_1$ and $X_2$ to represent the identities of the persons from two input videos.The idea is that if the two videos contain the same person (i.e.  $X_1$ = $X_2$) the distance between the feature vectors
should be small.  Otherwise, the distance should be large if
the persons are different (i.e. $X_1$ \(\neq\) $X_2$ ). Similar to ~\cite{mclaughlin16_cvpr}, we also use another loss (i.e. identity loss \textit{Loss}$_{id}$) to each branch of the Siamese network to predict
the person\textup{'}s  identity.  We use a linear classifier to predict one of the person\textup{'}s  identity from the feature vector extracted through each branch of the Siamese network. We then apply a Softmax loss over the prediction for each Siamese branch.
The final loss is the combination of two identity losses (\textit{Loss}$_{id1}$ and \textit{Loss}$_{id2}$) from each Siamese branch and the hinge loss as follows:\newline

\hspace{2cm}\(\mathbb{L}\)$_{final}$ = \(\mathbb{L}\)$_{id1}$+\(\mathbb{L}\)$_{hinge}$+\(\mathbb{L}\)$_{id2}$\hspace{2cm}(8)\newline

We use stochastic gradient decent to optimize the loss
function define in Eq. 8.  After training, we remove all loss
functions including the identity and hinge losses.   During
testing, we only use the feature vectors to compute the distance between two input videos for re-identification.

\begin{figure*}
\includegraphics[width = 0.85 \linewidth]{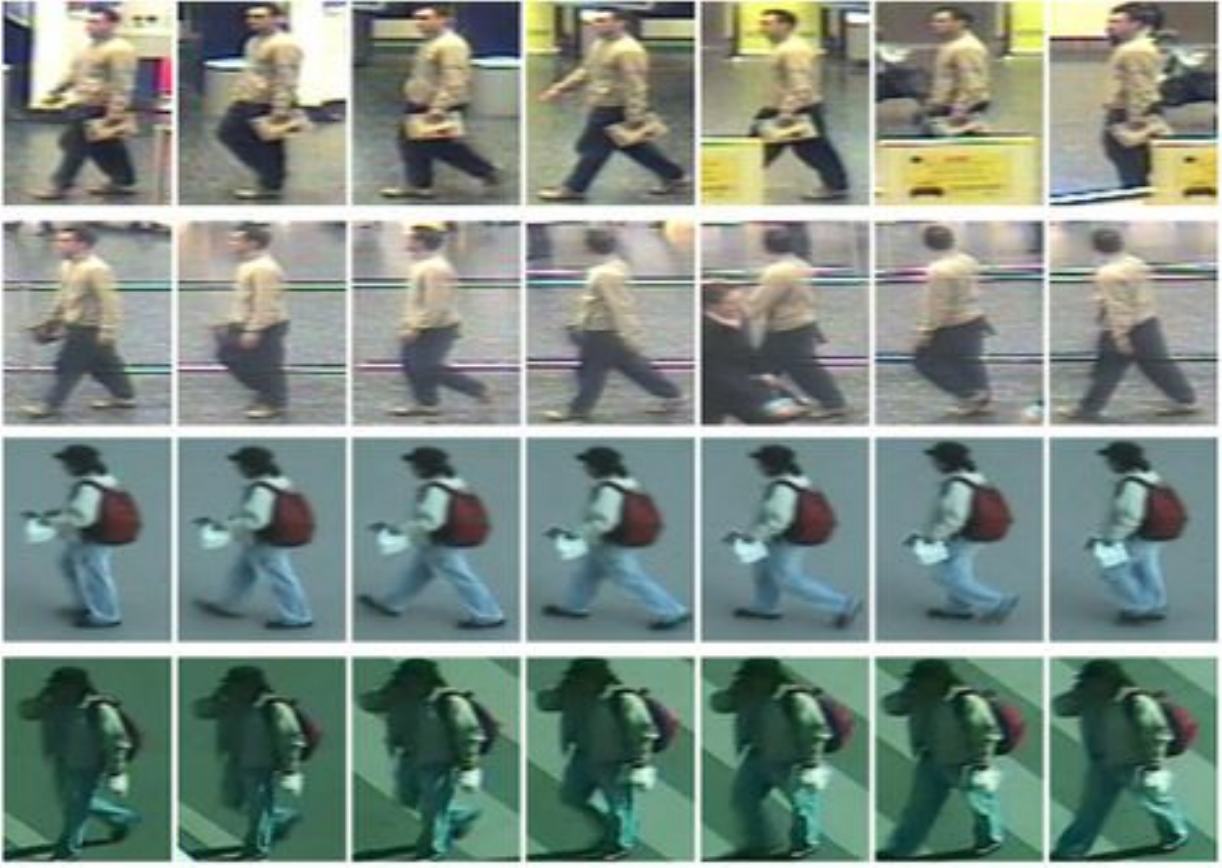}
\caption{Sample images of two benchmark datasets used in our
experiments.   The  first  two  rows  show  sample  images  from  the
iLIDS-VID dataset captured by two different cameras.  The next
two rows show sample images from the PRID-2011 dataset.}
\end{figure*}

\section{Experiments}
In this section, we firstly introduce the datasets used in
our experiments (Sec. 4.1).   We then describe the experimental setup and some implementation details (Sec. 4.2).
We present the results of experiment in Sec 4.3 and Sec 4.4.
\subsection{Datasets}
We  conduct  experiments  on  two  benchmark  datasets:
iLIDS-VID ~\cite{wang14_eccv} and PRID-2011 ~\cite{hirzer11_scia}. 

\textbf{iLIDS-VID Dataset:}
This dataset consists of video sequences of 300 persons where each person is captured by a pair of non-overlapping cameras.  The length of each video sequence varies from 23 to 192 frames with an average of 73 frames. The dataset is quite challenging due to lot of occlusions, illumination changes, background clutters and so
on.\newline
\textbf{PRID-2011  Dataset:
}This  dataset  contains  video  sequences of 749 persons.  For the first 200 persons (or identities), there are two video sequences captured by two different cameras.  The remaining persons appear in only one camera.  Each sequence contains between 5 to 675 frames,
with an average of 100 frames. In terms of complexity this
dataset is relatively simple than iLIDS-VID.

Some sample frames of these three datasets are shown in
Figure 5. Table 1 shows the summary of these two benchmark datasets.

\begin{table}[h]
  \caption{Summary of basic information of the two datasets used
in our experiments.}
  \label{tab:freq}
  \begin{tabular}{ccl}
    \toprule
    Dataset&iLIDS-VID&PRID-2011\\
    \midrule
    Total no. of id. & 300& 749\\
    No. id in multiple cameras & 300& 200\\
    No. track-lets & 600 & 400\\
    No. of boxes & 44k& 40k\\
    Image resolution & 64x128& 64x128\\
    No. of camera & 2& 2\\
    Detection procedure & hand& hand\\
    Complexity & Challenging& Simple\\
  \bottomrule
\end{tabular}
\end{table}

\subsection{ Setup and Implementation Details}
We follow the experiment protocol of McLaughlin
et al.~\cite{mclaughlin16_cvpr}.  On each of the two datasets (iLIDS-VID and PRID-
2011), we randomly split the dataset into two equal subsets
where one subset is used for training and remaining one for
testing.   For evaluating  our proposed  method,  we use  the
Cumulative Matching Characteristics (CMC) curve which
is a ranking based evaluation metric.  In the ideal case, the
ground-truth video sequence should have the highest rank.
For each dataset, we repeat the experiment 10 times and report the average result over these 10 runs.  In each run, we
randomly split the dataset into training/test sets.  
Standard
data  augmentation  techniques,  such  as  cropping  and  mirroring, are applied to increase the amount of training data.
We initialize the weights in the network using the initialization technique in ~\cite{he15_iccv}. For training our network, we consider
equal numbers of positive and negative samples. We set the
margin in the hinge loss (Eq. 7) as m = 2.  The network
is trained for 1100 epochs with a batch size of one.   The
learning rate in the stochastic gradient descent is set to be 1$e^{-4}$. Due to
the variable-length of video sequences in both datasets, we
use sub-sequences of 16 consecutive frames (
T
= 16
) during training. Sometimes, this length is greater than the real
sequence length.  In that case, we consider the whole set of
images (frames) as the sub-sequence. A full epoch consists of a pair of positive and negative sample. 
 During testing,
we consider a video sequence captured by the first camera
as the probe sequence and a video sequence captured by the
second camera as a gallery sequence.  We use at most 128
frames in a testing video sequence.  Again, if the length is
greater than the real sequence, we consider the whole set of
images as the video sequence. Similar strategies have been
used in previous work ~\cite{mclaughlin16_cvpr}.

\subsection{Results}
We present the results on the two benchmark datasets
and compare with other state-of-the-art methods in Table 2
and Table 3. From  the  CMC  rank,  we  see  that
our method outperforms all other state-of-the-art methods by nearly 4\(\%\) and 11\(\%\) on rank-1 accuracy on the iLIDS-VID and PRID-2011 dataset,respectively. Figure 6 shows some qualitative retrieval results after applying  our  proposed  method  on  the  challenging  iLIDS-VID dataset. We also show some failure cases in Figure 7. 

\begin{table}[h]
  \caption{Comparison of our proposed approach with other state-
of-the-art methods on the iLIDS-VID dataset in terms of CMC(\(\%\))
at different ranks.}
  \label{tab:freq}
  \begin{tabular}{cclll}
    \toprule
    Dataset&iLIDS-VID\\
    \midrule
    Method & Rank-1& Rank-5& Rank-10& Rank-20\\
   \textbf{ Ours} & \textbf{66}& \textbf{90}& \textbf{95}& \textbf{99}\\
    Xu et al.[26] & 62& 86& 94& 98\\
    Zhou et al.[35] & 55.2& 86.5& -& 97\\
    McLaughlin et al.[18] &58&84& 91& 96\\
    Yan et al.[27] & 49.3& 76.8& 85.3& 90.1\\
    STA[14] & 44.3&71.7&83.7 & 91.7\\
    VR[23] &35& 57& 68&78\\
    SRID[8]& 25& 45& 56& 66\\
    AFDA[10] &38& 63&73& 82\\
    DTDL[7] & 26&48& 57& 69\\
    
  \bottomrule
\end{tabular}
\end{table}
\begin{table}[h]
  \caption{Comparison of our proposed approach with other state-
of-the-art methods on the PRID-2011 dataset in terms of CMC(\(\%\))
at different ranks.}
  \label{tab:freq}
  \begin{tabular}{cclll}
    \toprule
    Dataset&PRID-2011\\
    \midrule
    Method & Rank-1& Rank-5& Rank-10& Rank-20\\
   \textbf{ Ours} & \textbf{88}& \textbf{97}& \textbf{99}& \textbf{99}\\
    Xu et al.[26] & 77& 95& 99& 99\\
    Zhou et al.[35] & 79.4& 94.4& -& 99.3\\
    McLaughlin et al.[18] &70&90& 95& 97\\
    Yan et al.[27] & 58.2& 85.8& 93.7& 98.4\\
    STA[14] & 64.1&87.3&89.9 & 92\\
    VR[23] &42& 65&78&89\\
    SRID[8]& 35& 59& 70& 80\\
    AFDA[10] &43& 73&85& 92\\
    DTDL[7] & 41&70& 78& 86\\
    
  \bottomrule
\end{tabular}
\end{table} 

\begin{figure*}
\includegraphics[width = 0.91 \linewidth]{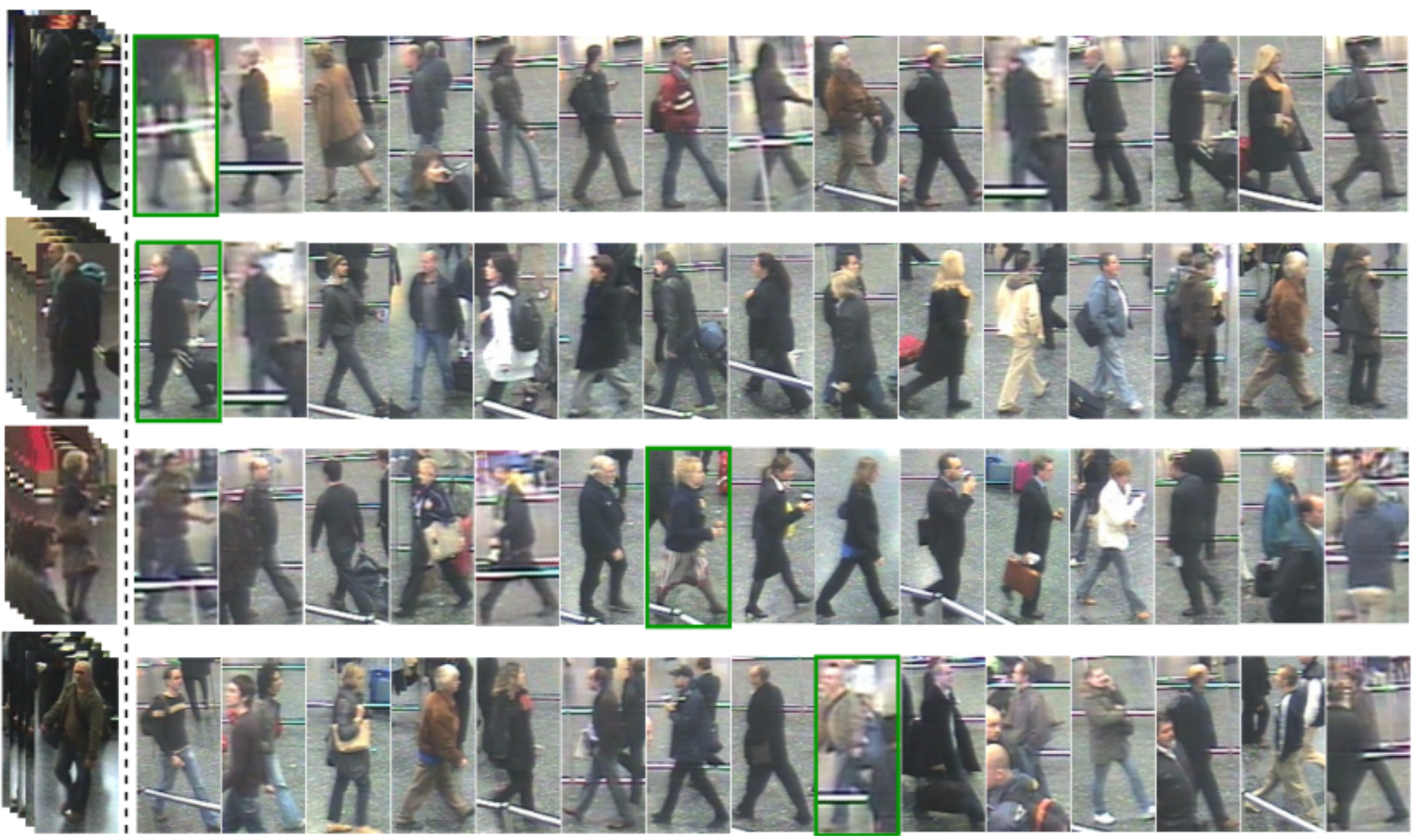}
\caption{Qualitative retrieval results of our proposed method on the challenging iLIDS-VID dataset. The first column represents the probe
video sequence.  The remaining columns correspond to retrieved video sequences sorted by their distances to the probe video sequence.
Here, we use a single image to represent each retrieved video sequence.  The green boxes indicate the ground-truth matches.  We can see
that the ground-truth matches are ranked very high in the list.}
\end{figure*}

\begin{figure*}
\includegraphics[width = 0.91 \linewidth]{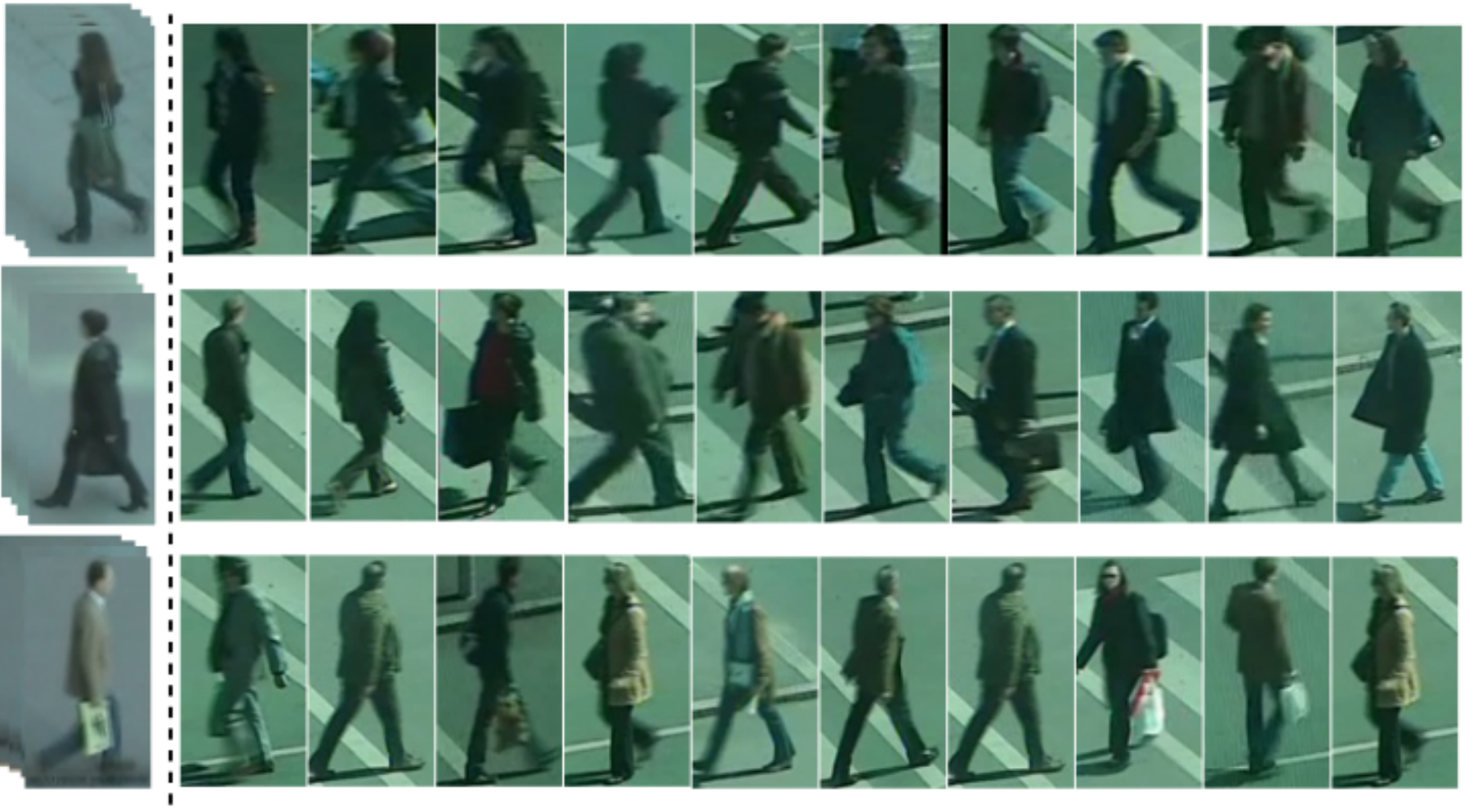}
\caption{Examples of some failure case of our proposed method. The first row indicates the probe sequence where single image in second
row represents retrieve gallery sequence of corresponding person}
\end{figure*}

\subsection{ Effect of Multiple Attention}
We  conduct  empirical  study  on  the  training  set  of  the
iLIDS-VID and PRID-2011 dataset to analyze the effect of the
multiple attention on the overall performance of the proposed network. We train the model on the
training videos and report the performance (CMC\(\%\)) on
the validation set for different number of attention layers in Table 4 and Table 5 respectively.  \\~\\
We observe that the performance gradually improves until number of attention layer is 3.  After that, the performance starts to drop.  Based on this empirical result, we choose number of attention layers as 3 in our experiments.It can also be observed from Table 4 and Table 5 that when the number of attention layers is 0, i.e. when we do not incorporate the concept of spatial attention the accuracies fall by a huge margin.
\begin{table}[h]
  \caption{ Performance for different number of attention layers
on the iLIDS-VID dataset.  Again, we report the performance in
terms of CMC \(\%\)}
  \label{tab:freq}
  \begin{tabular}{cclll}
    \toprule
    Dataset&iLIDS-VID\\
    \midrule
    No. of Att. layers & Rank-1& Rank-5& Rank-10& Rank-20\\
   3& 66& 90& 95& 99\\
    2 & 64& 88& 95& 99\\
   1 &63 & 88&95 &98 \\
	 0 &60 & 88&94 &98 \\
    
  \bottomrule
\end{tabular}
\end{table}
\begin{table}[h]
  \caption{ Performance for different number of attention layers
on the PRID-2011 dataset.  Again, we report the performance in
terms of CMC \(\%\)}
  \label{tab:freq}
  \begin{tabular}{cclll}
    \toprule
    Dataset&PRID-2011\\
    \midrule
    No. of Att. layers & Rank-1& Rank-5& Rank-10& Rank-20\\
   3& 88& 97& 99& 99\\
    2 & 84& 96& 99&99 \\
   1 & 80& 95& 98& 99 \\
    	 0 &80 & 96&99 &99 \\

  \bottomrule
\end{tabular}
\end{table}
\

\section{Conclusions}
In this paper, we have proposed an attention-based deep
architecture  for  video-based  re-identification.   The  attention module calculates frame-level attention scores, where
the attention score indicates the importance of a particular
frame.  We perform
experiments on two benchmark datasets and compare with
other state-of-the-art approaches.  We demonstrate that our
proposed method outperforms to other state-of-the-art approaches.\\~\\\\~\\


\bibliographystyle{ACM-Reference-Format}
\bibliography{sample-bibliography}

\end{document}